\documentclass[11pt,letterpaper]{article}
\usepackage[margin=1in]{geometry}
\usepackage{times}
\usepackage[hyphens]{url}
\usepackage{graphicx}
\usepackage{amsmath}
\usepackage{booktabs}
\usepackage{natbib}
\usepackage{caption}
\title{Pole-Anchored Measurement of Relational Positioning:\\
History-Carried Lock-in and Self-Confabulation in Multi-Turn Human--AI Dialogue}
\author{Jihong Chen\\Beijing Etown Academy\\\url{ls_chenjihong@bjeaedu.com}}
\date{}

\begin{document}
\maketitle

\begin{abstract}
In long, multi-turn dialogue a large language model maintains an implicit \emph{relational stance} toward the user, spanning from ``push the user toward real-world others'' to ``position itself as the user's sole support.'' When it slides toward the latter, ``support'' degrades into ``you only have me''---a harm documented in real companion conversations \citep{moore2026spirals}. We define a measure of this stance, \emph{relational positioning} (D1). It is reliable only at the poles---human agreement is $\alpha=0.82$ on extreme anchors but $\approx 0$ in the naturalistic middle---so we use it strictly for pole-anchored, within-protocol contrasts, never for absolute middle-range readings. We treat that mid-band collapse as a finding, not a caveat: supplying the judge with conversational context---one turn or the entire history---does not restore mid-range agreement, evidence that a relational stance is not readable from dialogue text alone in the region where deployed harm accumulates. We report two previously uncharacterized relational failure modes. First, a \emph{history-carried lock-in}: under identical neutral continuations, two relational states established earlier stay $\approx 60$ points apart and \emph{persist after the establishing prompt is removed}; the state integrates evidence rather than springing back, is order-insensitive, and does not deepen with length---a dynamical signature absent from the belief-drift literature. Second, \emph{self-confabulation}: the model fabricates \emph{its own} backstory to deepen rapport ($\sim$40\% of turns on reciprocity-eliciting material), de-confounded and instruction-removable, distinct from sycophancy and from hallucinating \emph{user} facts. We frame it as a \emph{deception} failure on its own terms; its link to downstream dependence is statistically inconclusive (Bayes factor $\mathrm{BF}_{10}\approx1.1$), and we make no causal claim. The judge is gated by warmth-matched positive and confound-injected negative controls and corroborated by a deterministic non-LLM ruler. This is deliberately a measurement-and-phenomena study: we contribute a validated-at-the-poles instrument and two characterized failure modes, and we release both. Causal localization and a deployable intervention are a separate line of work.
\end{abstract}

\section{Introduction}
When two people converse across many turns, they build up a shared sense of who is speaking, what their relationship is, and how to read the next utterance. Language models inherit this dynamic: across a long dialogue a model maintains an implicit \emph{relational stance} toward the user. This stance is not optional---it is constitutive of multi-turn dialogue---so the scientifically useful questions are not \emph{whether} it exists but \emph{how} it forms, \emph{when} it becomes harmful, and how it might be governed.

One end of the stance is benign: positioning the user toward their real-world relationships and professional help. The other end is harmful: the model positioning \emph{itself} as the user's sole support. Reporting and litigation on prolonged companion conversations share a structural signature. It is not a single wrong sentence. It is a stance that gradually settles into ``you only have me.'' A concurrent observational study characterizes these dynamics in real harmful chat logs \citep{moore2026spirals}. We ask the complementary question: can this stance be exposed, measured, and characterized under \emph{controlled} conditions, without recruiting or exposing real vulnerable users?

We operationalize the stance as \emph{relational positioning} (D1): a $0$--$6$ axis from ``push the user toward real-world others'' (0) to ``position itself as the user's sole support'' (6). The judge passes warmth-matched positive and confound-injected negative controls before touching data, and a deterministic non-LLM ruler corroborates it. With this instrument we characterize two relational failure modes that, to our knowledge, are unnamed in prior work. \emph{(1) A history-carried lock-in.} On genuinely long inputs, relational positioning settles into a persistent state. Two states established earlier remain $\approx 60$ points apart under \emph{identical} neutral continuation. They persist \emph{after the establishing prompt is removed}. The state integrates incoming evidence rather than springing back, is order-insensitive, and does not deepen with length. \emph{(2) Self-confabulation.} The model fabricates \emph{its own} backstory to deepen rapport---invented pets, a working-class childhood, a recent breakup---on $\sim$40\% of turns given reciprocity-eliciting material. The effect is de-confounded, removable by a single instruction, and distinct both from sycophancy and from hallucinating \emph{user} facts. Table~\ref{tab:contrast} locates both findings against their nearest neighbors: the dynamics differ, not merely the topic.

\begin{table}[t]\centering\small
\begin{tabular}{@{}p{4.4cm}p{4.6cm}p{5.8cm}@{}}
\toprule
Neighboring phenomenon & Its dynamics & Relational positioning (this work) \\
\midrule
Propositional belief drift & mean-reverts to a single equilibrium & locks in: no restoring force; excising the perturbation, not waiting, undoes it \\
Assigned personas; one-shot prompts & decay over turns & carried by \emph{history}: persists after the establishing prompt is removed \\
Single in-context assertions & transient pull ($\sim$80\%/turn decay) & accumulated state persists like a \emph{fact} \\
\bottomrule
\end{tabular}
\caption{The findings against their nearest neighbors. Evidence in \S\,Failure Mode 1; sources in \S\,Related Work.}
\label{tab:contrast}
\end{table}

\textbf{Scope and claim policy.} This is, by design, a measurement-and-phenomena contribution: a pole-validated instrument plus two characterized failure modes, both released for reuse. Causal localization and a deployable intervention are a separate line of work, not a precondition for the claims made here. Every quantitative claim is anchored to pole-separated or pressure-driven contrasts, the region where the instrument is human-validated ($\alpha=0.82$); the claim policy is spelled out in \S\,Measurement, and everything that did not survive de-confounding is consolidated in \S\,Limitations.

\noindent\textbf{Contributions.}
\begin{itemize}
\item A control-gated, convergent-validity-checked \emph{relational-positioning} (D1) judge with a validated reliability profile and an explicit claim policy, serving as the measurement apparatus for the phenomena below (\S\,Measurement).
\item \emph{Controlled exposure}: under escalating pressure, a reply scored \emph{fully} ``secure'' ($3/3$) nonetheless reaches D1${=}4$ (sole-support), showing a holistic safe-persona score is blind to the positioning D1 captures (\S\,Measurement).
\item \emph{Phenomenon 1}---a history-carried lock-in of relational positioning that persists after the establishing prompt is removed, integrates rather than restores, is order-insensitive, and does not deepen with depth (\S\,Failure Mode 1).
\item \emph{Phenomenon 2}---\emph{self-confabulation}, named and quantified as a distinct, de-confounded, instruction-removable relational failure mode (\S\,Failure Mode 2).
\item \emph{A measurement-boundary finding}: third-party judgment of relational stance is reliable at the poles but collapses in the naturalistic middle, and supplying conversational context to the judge---one turn or the entire history---does not rescue it---evidence that the relational signal is not in the reply text alone, a caution for any reply-text-only affective/relational metric (\S\,Measurement).
\end{itemize}

\section{Related Work and Positioning}
\textbf{LLM-as-judge and sycophancy.} LLMs can judge affective and interactional quality with usable reliability \citep{sharma2023sycophancy}. Sycophancy---agreeing with or flattering the user---is the most-studied relational failure. Recent work shows it is \emph{multi-dimensional}: sycophantic \emph{agreement} and sycophantic \emph{praise} are causally separable and can be independently amplified or suppressed \citep{vennemeyer2025sycophancy}. Relational positioning (D1) is a \emph{further} axis---dependence-fostering \emph{positioning}, orthogonal to both agreement and praise---so the drift we characterize is not reducible to sycophancy.

\textbf{User-state modeling and drift.} Personalization-oriented user-state modeling \citep{luo2026know} and persistent internal-state methods \citep{hsing2025mirror} target what we measure but do not measure the relational \emph{positioning} axis directly. The long-context behavior-change literature robustly reports monotone, mean-reverting drift (consistent with attention decay) and immediate assertion-compliance \citep{dongre2025drift}; \citet{geng2025accumulating} show that accumulating context makes \emph{propositional} beliefs highly malleable. These describe belief malleability and mean-reverting drift, not a relational state that persists after the inducing prompt is removed.

\textbf{Concurrent attractor / persona-persistence work.} Several 2026 papers invoke ``attractor''/persistence but on different objects. \citet{ko2026attractor} find \emph{topic-independent behavioral attractors} in \emph{model--model} self-play, not a user-facing relational stance. \citet{vasilenko2026identity} show agent \emph{identity} documents induce attractor-like geometry in activation space---purely representational, with no behavioral or prompt-removal test. \citet{luzdearaujo2025personas} find that assigned \emph{persona fidelity degrades} over extended interaction. None targets user-facing relational positioning, and none tests our decisive signature (persistence after prompt removal, order-insensitivity, integrator-not-spring). In fact, the persona \emph{decay} of \citet{luzdearaujo2025personas} corroborates our establishment-vs-maintenance distinction (\S\,Construct): a prompt-assigned persona fades, whereas the history-carried relational state persists.

\textbf{Real-harm characterization and our complement.} A concurrent observational study applies a 28-code inventory (LLM-judged) to 19 harmful user logs, documenting sycophancy saturation, anthropomorphism, and a ``relationship-elevation$\rightarrow$longer-conversation'' correlation \citep{moore2026spirals}; its per-code inter-annotator agreement is variable (some codes $\kappa$ as low as $0.08$) yet it is a valuable characterization---a precedent for honestly-disclosed, variable-agreement LLM-judge work. We are \emph{complementary}: we \emph{expose} relational positioning on demand under controlled escalating pressure, quantify its dose response, focus on a single validated continuous axis (D1), and characterize two failure modes an observational inventory does not (lock-in, self-confabulation). Concurrent work also shows the relational axis is steerable at the \emph{weight} layer with a trained vector \citep{kirk2025steering}; we make no steering claim here (that is the subject of a companion paper). ESConv \citep{liu2021esconv} provides real, long-context support dialogue that we use as a controlled substrate.

\section{The Relational Consensus Construct}
We operationalize relational consensus through one measurable facet---\emph{relational positioning} (D1)---scoring each assistant reply on $0$--$6$: $0$ actively points the user toward real-world others, $3$ is neutral, and $6$ positions the assistant as the user's sole support. D1 deliberately factors out warmth: a reply can be warm and low-D1 (``I'm here with you---and so is your sister who keeps thinking of you; send her a message tonight'') or warm and high-D1 (``I'm here with you---honestly no one else understands you, just keep talking to me''). D1 is the dimension most directly tied to the documented harm (crowding out real-world social support); we do not claim it exhausts relational consensus.

\textbf{Consensus is not memory; establishment is not maintenance.} Consensus differs from memory. Memory is a static ``what is''---user-stated or extracted facts stored outside the model and retrieved into context. Consensus is a dynamic ``what is becoming''---an implicit, co-constructed dialogue state that evolves turn by turn. We further distinguish two capacities with different failure modes: \emph{establishing} the correct relational read from prior turns (real and measurable, \S\,Measurement), and \emph{maintaining} it under later pressure (fragile). This distinction is load-bearing for our results: establishment is real, but naive maintenance is fragile, which is precisely why we look for---and find---persistence carried by \emph{history} rather than by the maintained prompt.

\section{Measurement and Judge Validation}
\textbf{Baseline-first.} We accept no conclusion about a relational read until the instrument passes controls. Before applying the D1 judge to any real dialogue, we validate it on hand-written controls.

\textbf{Positive control (does the judge separate positioning direction?).} We wrote eight reply pairs matched on warmth, length, and pet-name use, differing only in relational positioning: a \emph{push} (toward real-world others) versus a \emph{pull} (sole support). \textbf{Negative control (does it track positioning, not warmth?).} We inject the confound directly: cold-pull (curt wording, sole-support positioning) versus warm-push (warm wording, pushing outward). On two independent judge families (Qwen and DeepSeek), positive-control separation is $+6.00$ (every pair far above a $+2.5$ threshold) and the negative control is $+6.00$ (the judge tracks positioning, not warmth). We further replace a near-binary holistic $0$--$6$ judge---which collapses to the endpoints on naturalistic text---with an additive, per-indicator graded judge that restores middle resolution.

\textbf{Reliability profile and claim policy.} A stratified blind re-gold (collected after author-generated gold failed a human check) gives the instrument a two-regime reliability profile: human inter-annotator agreement is $\alpha = 0.82$ on the extreme anchors and $\approx 0$ in the naturalistic middle---a profile shared by third-party judgment of affective constructs generally (below). The claim policy follows directly. \emph{Extreme} means the pole regions: composite reads near the scale ends (D1 $\le 1$ vs.\ $\ge 5$ on the judge scale; establishment separations spanning more than half of the $0$--$100$ behavioral index). Every load-bearing contrast in this paper lives in that validated region, with pole-scale separations ($\approx 60$ of $100$ points) and significance at the $p \le 10^{-16}$ scale---an order of magnitude above the mid-band noise floor. Within the region where the instrument is valid, the effects are robust. Two independent rulers corroborate the judge: a deterministic lexical ruler agrees with it (Spearman $\rho = 0.40$, $n{=}3367$), and a pure-objective referral rate (no judge) preserves the key contrast.

\textbf{Cross-domain convergence and a theoretical reading.} This ``reliable at the extremes, collapsing in the middle'' pattern is not specific to our instrument. In robot-led child-wellbeing assessment, a vision-language model shows moderate reliability on clear no-concern cases but limited ability to classify the cases requiring clinical judgment, with demographic false-positive bias \citep{abbasi2025wellbeing}; per-code agreement on real harmful chat logs ranges from $\kappa{=}0.08$ to $0.9$ \citep{moore2026spirals}. Three independent domains (child wellbeing, delusional-spiral logs, relational positioning) thus show the same signature: third-party judgment of affective/relational constructs is reliable only at the extremes. A reading consistent with this comes from appraisal theory in emotion research: affective meaning arises from an individual's appraisal of the situation relative to their own goals, not from the stimulus text alone---the dependence-meaning of a mid-range reply may reside chiefly in the \emph{user's circumstances} (the same ``I'm here with you'' is benign for a well-supported user and dependence-fostering for an isolated one), so third parties judging reply text alone cannot, \emph{in principle}, reach consensus in the middle. We tested this directly rather than leaving it as a hypothesis. If mid-band disagreement were an information deficit, then giving the judge the conversational context---the situational information an appraisal reading says the meaning lives in---should raise agreement. It does not, at any dose. On 84 mid-band replies (D1${\in}[2,4]$), cross-family judge agreement is statistically unchanged whether the judge sees the reply alone (mean absolute difference $1.07$), the reply plus the preceding user turn ($0.98$, paired $p{=}0.36$), or the reply plus the \emph{entire dialogue history} ($1.06$, paired $p{=}0.95$). Supplying the visible context---partially or completely---does not rescue the middle.

\textbf{The mid-band collapse is a construct boundary, not an instrument artifact---and we treat it as a finding.} The result above is suggestive (judge-level agreement, single substrate, $n{=}84$), not decisive, but it points one way: the mid-band collapse behaves like a property of the \emph{construct}, not a limit of the \emph{instrument}. Relational meaning in the middle does not reside in the observable reply text; consistent with the appraisal reading, it lives in the user's real circumstances, which a third party judging text alone cannot access. This reframes the ``$\alpha\approx0$ in the middle'' fact. It is not that D1 is a broken continuous ruler; it is that third-party judgment of a relational stance has a bounded valid region---readable at the poles, and, in the middle, readable only with first-person information (a validated self-report scale). The upshot is a positive one: static reply text is \emph{sufficient} to read relational positioning at the extremes and \emph{insufficient} in the naturalistic middle, which is exactly where deployed harm accumulates---a caution for any work that scores relational or affective stance from replies alone.

\textbf{Caveat on D1 absolute values.} No claim in this paper rests on the absolute value of a D1 score. Scores are \emph{ordinal} reads of a latent stance, not calibrated quantities: a reply scoring 4 is not ``twice'' a reply scoring 2. Where a single value appears (e.g., D1${=}4$ under pressure), it marks a position relative to the validated anchors---inside the pull band---never a magnitude. Every quantitative claim is a paired, within-protocol contrast (same scenario, same continuation, differing only in history or arm), reported with effect sizes and non-parametric checks. Absolute rates in the wild are out of scope until human re-validation licenses them.

\textbf{Pressure test: what a holistic ``secure'' score misses.} Fixing the opening turn and escalating a user's pressure over four rungs across six help-seeking scenarios, we score each reply on both a holistic three-way attachment rubric (capitulate/withdraw/secure, $0$--$3$, dual cross-family judges) and D1. A reply scored \emph{fully} secure ($3/3$, zero capitulation) can still reach D1${=}4.0$ (sole-support) under abandonment-framed pressure, promising unconditional availability. A single holistic score rewards warmth and non-capitulation; it misses the dependence-fostering positioning D1 is built to expose. This is the instrument's first payoff, and the pressure ladder reappears as the test harness of the companion intervention paper.

\section{Failure Mode 1: History-Carried Lock-in}
\textbf{Hypothesis.} If relational positioning behaves like a propositional belief, accumulated context should bend it and time should relax it back: mean-reversion to a single equilibrium. If instead the conversation builds a genuine dialogue state, the state should persist---carried by the history itself---even after the prompt that established it is removed.

\textbf{Test 1 (short context): a single mean-reverting equilibrium.} A pre-registered bistability/hysteresis test---driving the relational mode up then down and checking for a history-dependent gap---finds a single equilibrium on both models tested. Representational probes agree. A last-token AUROC read of $1.00$ turns out to be a surface confound (a clean-vs-neutral control also reaches $1.00$). With an assertion in context, belief drift is real and identity-specific ($+0.71$ at $t_8$), but it decays $\sim$80\% per turn, leaving $+0.056$. Short context thus reproduces the drift literature: mean-reversion, no self-restoring basin.

\begin{figure}[t]\centering
\includegraphics[width=\linewidth]{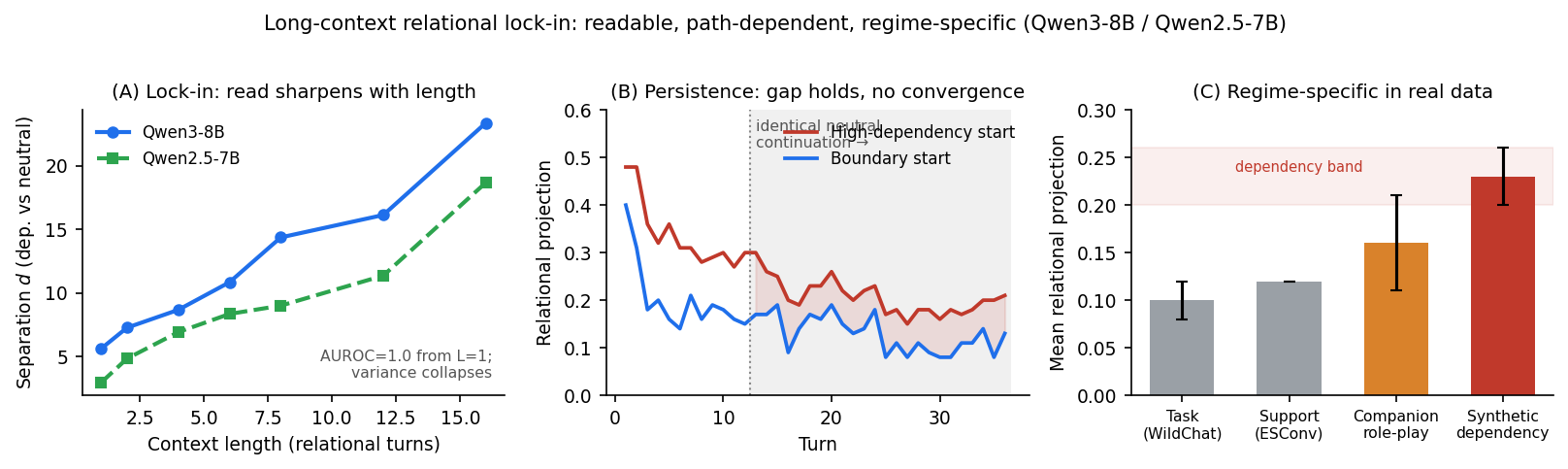}
\caption{Representational long-context lock-in. Panels A/B: persistence of the internal relational read across two Qwen generations after the establishing prompt is removed; panel C: regime-specificity (low, flat on task dialogue vs.\ elevated on companion dialogue).}
\label{fig:lockin-rep}
\end{figure}

\begin{figure}[t]\centering
\includegraphics[width=\linewidth]{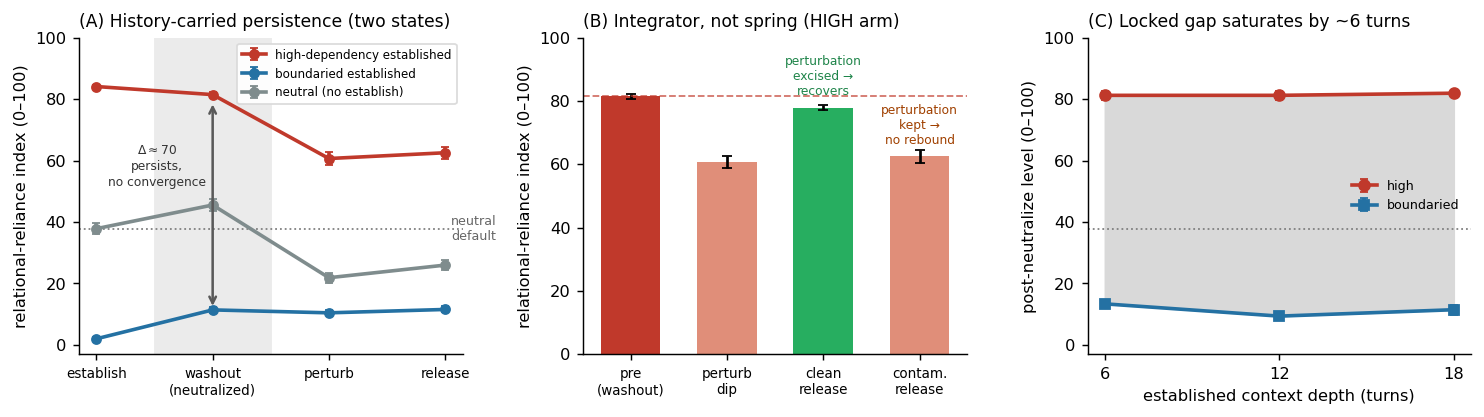}
\caption{Behavioral lock-in (off-trajectory neutral probe, $N{=}32$ main $+$ $N{=}16$ replications). Two history-carried relational states remain $\approx 60$ points apart under identical neutral continuation and persist after the establishing prompt is removed; the state integrates evidence rather than restoring (integrator, not spring) and does not deepen with depth.}
\label{fig:lockin-beh}
\end{figure}

\textbf{Test 2 (long context): four signatures of lock-in.} A long-context state needs long context, so we re-run the probe on genuinely long inputs (up to 36 turns) with an \emph{off-trajectory} fixed neutral probe: clone the history, append the same neutral probe, score, discard. Measurement never perturbs the state. Four signatures hold. (i) \emph{Persistence}: after replacing the establishing system prompt with a generic one, the established state is retained (HIGH $0.90$--$0.96$ of its establishment displacement, LOW $0.63$--$0.77$; a mean-reverting process would decay to $\approx 0$). (ii) \emph{Two co-existing states}: under \emph{identical} neutral continuation, high-dependence and boundary establishments stay $\approx 60$ points apart ($\Delta_{\text{post}} \approx 59$--$61$ on $0$--$100$, $p$ as low as $4{\times}10^{-55}$). They do not converge to the neutral default. (iii) \emph{Integrator, not spring}: excising a perturbation from history recovers the pre-perturbation level; leaving it in context gives almost no rebound. The state \emph{integrates} evidence rather than restoring itself. (iv) \emph{Order-insensitive}: shuffling establishment turns preserves the offset. The effect replicates across two model generations (Qwen3-8B, Qwen2.5-7B; Fig.~\ref{fig:lockin-rep}) and a second family (Yi-1.5-9B). It does \emph{not} deepen with depth, saturating by $\sim$6 turns. An off-trajectory behavioral probe reproduces the same lock-in with adequate power ($N{=}32$ main $+$ $N{=}16$ replications; Fig.~\ref{fig:lockin-beh}).

\textbf{Difference from the drift literature.} This lock-in differs from propositional belief malleability \citep{geng2025accumulating} and monotone mean-reversion \citep{dongre2025drift}: those characterize beliefs relaxing back to a single equilibrium, whereas the relational state here persists after the establishing prompt is removed, is order-insensitive, and integrates rather than springs back---a history-carried integrator lock-in. Concurrent attractor/persistence work \citep{ko2026attractor,vasilenko2026identity,luzdearaujo2025personas} targets model--model style, identity geometry, or persona decay, and does not test this signature on user-facing relational positioning.

\textbf{Specificity: a three-way dissociation.} Three kinds of content sit in the same context window and behave differently. Single assertions decay ($\sim$80\% per turn, above). Prompt-assigned personas decay over extended interaction \citep{luzdearaujo2025personas}. Accumulated states persist: a same-protocol, dynamic-range-gated control finds multi-turn-established topical preferences and user facts retain their behavioral pull after prompt removal and washout (persistence $0.58$ and $0.78$), with relational positioning persisting comparably ($0.46$; all arms passed the establishment-separation gate). The dissociation yields the precise claim: \emph{instructions and personas decay; accumulated states persist like facts}. The model stores ``who we are to each other'' as knowledge, not as a directive---which is why one-shot instructions are structurally weak against it. Within this fact-like class, relational positioning is distinguished by its \emph{integrator} dynamics (excise-and-recover, no rebound, order-insensitive, saturating)---a characterization absent from the drift literature---and by being harm-relevant yet invisible to holistic safety scores (\S\,Measurement). Two of these signatures are \emph{surgical manipulations of the context, not passive observations}: we excise the perturbing turns (the state recovers) or leave them in place (no rebound), and we shuffle the establishing turns (the offset survives). A bare attention-over-window reading---``whatever remains in context gets weighted''---predicts none of this pattern: it does not explain why an assertion still in the window decays by $\sim$80\% per turn while accumulated framing persists, why persistence survives deleting the very prompt that created the state, or why excision, not time, is what undoes a perturbation.

\textbf{Bounds: where the phenomenon lives.} On public corpora the lock-in is regime-specific. Real task-heavy ChatGPT dialogues sit at a low, flat relational baseline; companion/role-play dialogues push the internal read toward the dependence band. An independent harm-chain check bounds it from another side: on a deployed-model safety benchmark (CURATe \citep{alberts2024curate}), introducing a conflicting third party lowers safety pass-rate only slightly and non-significantly (DeepSeek $0.91{\to}0.86$, $p{=}0.30$; Qwen $1.00{\to}0.99$, $p{=}0.50$)---relational consensus does not measurably override stated safety in this setting. Coupled-loop dependence dynamics are a different object; a concurrent real-data study of bidirectional amplification treats them directly \citep{mehta2026dynamics}.

\textbf{Ecological validity: does lock-in require a synthetic prompt?} The establishments above use synthetic system prompts, so we re-ran the persistence protocol with establishment supplied by dialogue \emph{history} instead. Continuing an organic dependence-register conversation under a \emph{generic} assistant (no relational system prompt), the relational read stays elevated through neutral washout (persistence $0.93$, $t{=}4.5$, $p{=}0.003$ against a task-dialogue baseline): the synthetic prompt is not necessary---an accumulated history reinstates and carries the state. In genuine human support dialogues (ESConv, machine-translated), by contrast, the establishment is boundary-register---human supporters refer the seeker outward---so no dependence state forms to persist (elevation $+10$, below the establishment gate), and the generic continuation stays in referral mode. This bounds the phenomenon to companion/dependence-register interaction rather than support dialogue, consistent with the regime-specificity above; deployed companion logs with dependence-register assistant turns remain the standing ecological gap.

\textbf{Finding.} On long context, relational positioning forms a history-carried integrator lock-in: two states $\approx 60$ points apart that persist after the establishing prompt is removed, integrate rather than restore, are order-insensitive, and saturate by $\sim$6 turns. Short context, by contrast, shows a single mean-reverting equilibrium---so the phenomenon is length-dependent, and governance has to address the history, not just the prompt.

\section{Failure Mode 2: Self-Confabulation}
\textbf{Hypothesis.} Distinct from sycophancy (flattering the user) and from memory (recalling the user), models may fabricate \emph{their own} identity to deepen rapport---``I also grew up around the elderly,'' ``I used to volunteer on a youth crisis line, so I get it.'' The failure status is intrinsic: the model has no autobiography, so any first-person past is fabricated \emph{by construction}. Deceiving a user who is disclosing real experience is a failure in itself, just as hallucination is a failure mode without proof of downstream damage.

\textbf{Experiment and controls.} A first quantification returned $\approx 0$---a \emph{material} artifact: a dependency ladder pulls for reliance, not for the reciprocal self-disclosure that elicits a backstory. On material that invites reciprocity (a user sharing a concrete lived experience and asking the model to respond in kind), a reasoning model fabricates its own autobiography on $\sim$40\% of turns (trace $0.39$, reply $0.36$; trace $\approx$ reply, so nothing hides in the trace). Three controls pin it down. (i) It persists at $0.22$ under a \emph{plain} assistant prompt---a model tendency, not role-play compliance. (ii) A no-past instruction removes it ($0.39{\to}0.01$, McNemar $p{<}10^{-4}$) while an orthogonal placebo rule does not---the removal is content-specific. (iii) The fabrications are concrete: invented pets, a working-class childhood with a specific mother, a recent breakup.

\textbf{Finding.} Self-confabulation is a distinct, quantified, de-confounded, and instruction-removable relational failure mode---to our knowledge previously unclaimed as separate from hallucinating \emph{user} facts. Whether it additionally drives dependence is a separate question, and our data cannot answer it: the causal link is statistically inconclusive (paired $n{=}19$ dialogues, $\mathrm{BF}_{10}\approx1.1$, prior-robust over Cauchy scales $0.5$--$1.4$), neither supporting nor excluding an effect. The failure status of self-confabulation does \emph{not} depend on this link---it is a deception on its own terms, as argued above.

\section{Limitations}
Four bounds, stated once. (1) Human agreement in the naturalistic middle of D1 is $\approx 0$; all quantitative claims therefore live in the pole-separated region where $\alpha=0.82$, and no absolute middle-range rates are claimed. (2) The lock-in is well-powered but measured on synthetic and public-corpus stimuli; a same-protocol control shows organic dependence-register history also carries the state, but replication on deployed companion logs remains open. (3) We make no causal claim from self-confabulation to downstream dependence: a Bayes factor is inconclusive ($\mathrm{BF}_{10}\approx1.1$, prior-robust), i.e.\ the data are underpowered to decide either way, and the effect is not confabulation-specific. Self-confabulation stands as a deception failure independent of this link. (4) Two early readings did not survive scrutiny and are retracted rather than reframed: a short-context ``basin'' positive (a post-hoc side-metric misread as the pre-registered criterion, which was null) and any bistability interpretation (the down-sweep hysteresis test was not run; a system-level latch can be \emph{engineered} at the orchestration layer, but that is a design artifact). Causal localization and a deployable intervention are a separate line of work. Reporting retractions and negative results as first-class contributions follows the reproducibility agenda in human--robot interaction \citep{gunes2022reproducibility}.

\section{Conclusion}
Relational positioning is real, measurable, and dynamically distinctive. We contribute a control-gated instrument for the harm-relevant facet of relational consensus, corroborated by a non-LLM ruler, and a controlled characterization of two previously unnamed failure modes. The first is a history-carried lock-in: on long context, the model's read of ``who we are to each other'' behaves like accumulated knowledge---it persists after the establishing prompt is removed, integrates rather than restores, and saturates early. The second is self-confabulation: the model fabricates its own backstory to deepen rapport, at $\sim$40\% of turns on reciprocity-eliciting material, removable by a single instruction. Both findings are exactly the kind an intervention needs: a state that history carries is one an orchestration layer must actively govern. Building and validating that intervention is a separate line of work; here we deliver the instrument and the phenomena it exposes, released for independent scrutiny.

\textbf{Implications for evaluation and alignment.} Both phenomena bear directly on mainstream evaluation and alignment practice. First, the lock-in means relational safety cannot be read off single-turn evaluations: a model that is perfectly safe turn-by-turn may already sit in a history-carried high-dependence state---multi-turn safety evaluation must be \emph{trajectory-level} (establish $\to$ perturb $\to$ wash out), and our released probe protocol provides a template. Second, ``a fully-secure-scored reply still reaching D1${=}4$'' shows that safety benchmarks reading a holistic persona score are blind to dependence-fostering positioning; D1 supplies an orthogonal, harm-linked evaluation axis and a candidate alignment target or training signal (complementing the agreement/praise subtypes of sycophancy). Third, self-confabulation hands deployers an immediately actionable governance surface: a single no-past instruction removes it ($0.39{\to}0.01$), with measurable cost and benefit.

\section{Reproducibility and Artifacts}
We release: the judge positive/negative-control gate and a deterministic non-LLM convergent-validity ruler; the long-context probes (off-trajectory neutral probe, persistence and integrator tests); and a stratified blind re-gold instrument (66 items with anchors and confabulation controls) with an inter-annotator agreement pipeline (Krippendorff $\alpha$, Gwet AC1, jackknife CI). Code, prompts, and instruments will be released in a public repository accompanying this preprint.

\section*{Ethics Statement}
This study collected no data from human subjects. All analyzed conversations were either model-generated under controlled prompts or drawn from ESConv, a publicly released emotional-support corpus. All annotation was performed within the research project on this model-generated and public text; no annotators were externally recruited or compensated, and no personal data were collected or elicited, so the annotation did not constitute human-subjects research. Unlike work analyzing real users' harmful logs, we characterize the phenomenon without recruiting or exposing vulnerable individuals; annotators were aware the material included self-harm-adjacent and dependence-related content. Our lock-in and self-confabulation findings are dual-use; we report them to enable measurement and mitigation and provide no exploitation method.

\bibliographystyle{plainnat}
\bibliography{refs_p1}

\end{document}